\documentclass{article}
\usepackage{microtype}
\usepackage{graphicx}
\usepackage{subfigure}
\usepackage{booktabs} %
\usepackage{amsthm}
\usepackage{amssymb}
\usepackage{mathtools}
\usepackage{hyperref}
\usepackage[nameinlink,capitalize]{cleveref}
\hypersetup{colorlinks=true,linkcolor=blue,citecolor=blue,urlcolor=blue,pdfborder={0 0 0}}
\usepackage[normalem]{ulem} %
\usepackage{mathtools}

\usepackage[utf8]{inputenc} %
\usepackage[T1]{fontenc}    %
\usepackage{hyperref}       %
\usepackage{url}            %
\usepackage{booktabs}       %
\usepackage{amsfonts}       %
\usepackage{nicefrac}       %
\usepackage{microtype}      %
\usepackage{proof-at-the-end}  %
\usepackage{multirow}
\usepackage{lscape}         %

\usepackage{graphicx,wrapfig}
\usepackage{amsfonts}
\usepackage{url}
\usepackage{enumitem}
\usepackage{caption}
\usepackage{amsthm}
\usepackage{thmtools}
\usepackage{thm-restate}

\theoremstyle{definition}

\theoremstyle{remark}

\usepackage{pifont}%

\usepackage{enumitem}

\usepackage{color, colortbl}
\definecolor{Gray}{gray}{0.9}
\newcolumntype{g}{>{\columncolor{Gray}}c}

\definecolor{Gray}{gray}{0.93}
\newcolumntype{a}{>{\columncolor{Gray}}c}

\newcommand{\cS}{{\mathcal{S}}}

\newcommand{\bc}{\begin{center}}
\newcommand{\ec}{\end{center}}

\newcommand{\bdm}{\begin{displaymath}}
\newcommand{\edm}{\end{displaymath}}

\newcommand{\beq}{\begin{equation}}
\newcommand{\eeq}{\end{equation}}

\newcommand{\bfl}{\begin{flushleft}}
\newcommand{\efl}{\end{flushleft}}

\newcommand{\bt}{\begin{tabbing}}
\newcommand{\et}{\end{tabbing}}

\newcommand{\beqn}{\begin{align}}
\newcommand{\eeqn}{\end{align}}

\newcommand{\beqs}{\begin{align*}} %
\newcommand{\eeqs}{\end{align*}}  %

\newcommand{\norm}[1]{\left\|#1\right\|}

\newcommand{\etal}{\emph{et al.}}
\newcommand{\Ebb}{\mathbb{E}}

\usepackage{amsmath,amsfonts,bm}

\def\eqref#1{equation~\ref{#1}}

\def\1{\bm{1}}

\def\vx{{\bm{x}}}

\DeclareMathAlphabet{\mathsfit}{\encodingdefault}{\sfdefault}{m}{sl}
\SetMathAlphabet{\mathsfit}{bold}{\encodingdefault}{\sfdefault}{bx}{n}

\DeclareMathOperator*{\argmax}{arg\,max}
\DeclareMathOperator*{\argmin}{arg\,min}

\usepackage{hhline}
\usepackage{algorithm2e}
\usepackage{hyperref}
\usepackage{diagbox}

\usepackage[accepted]{icml2023}

\usepackage{epsfig}
\usepackage{graphicx}
\usepackage{amsmath}
\usepackage{amssymb}
\usepackage{bbm}
\usepackage{mathrsfs}
\usepackage{multirow}
\usepackage{bbm}
\usepackage{mathtools}
\usepackage{url}            %
\usepackage{booktabs}       %
\usepackage{amsfonts}       %
\usepackage{nicefrac}       %
\usepackage{threeparttable}
\usepackage{graphicx}
\usepackage{amsmath}
\usepackage{color}
\usepackage{subfigure}
\usepackage{multicol}
\usepackage{listings}
\usepackage{wrapfig}
\usepackage{amssymb}
\usepackage{pifont}
\usepackage[utf8]{inputenc}
\usepackage{subfigure}
\usepackage{xcolor}
\usepackage{comment}
\usepackage{amsmath,amssymb,amsfonts}
\usepackage[justification=centering]{caption}
\usepackage{makecell}
\usepackage{indentfirst}
\usepackage{shorttoc}
\usepackage{caption}
\usepackage{thmtools}
\usepackage{thm-restate}

\usepackage{cuted}
\usepackage{capt-of}

\definecolor{deepred}{rgb}{0.631,0.102,0.102}
\definecolor{skyblue}{HTML}{126da2}
\definecolor{mildyellow}{HTML}{FFF2CC}
\hypersetup{
     colorlinks = true,
     breaklinks = true,
     urlcolor = deepred,
     citecolor = blue
     }

\usepackage{authblk}
\makeatletter
\newcommand{\printfnsymbol}[1]{%
  \textsuperscript{\@fnsymbol{#1}}}
\makeatother

\newenvironment{packeditemize}{
\begin{list}{$\bullet$}{
\setlength{\labelwidth}{8pt}
\setlength{\itemsep}{0pt}
\setlength{\leftmargin}{\labelwidth}
\addtolength{\leftmargin}{\labelsep}
\setlength{\parindent}{0pt}
\setlength{\listparindent}{\parindent}
\setlength{\parsep}{0pt}
\setlength{\topsep}{3pt}}}{\end{list}}

\icmltitlerunning{
Revisiting Data-Free Knowledge Distillation with Poisoned Teachers
}

\begin{document}

\twocolumn[

\icmltitle{
Revisiting Data-Free Knowledge Distillation with Poisoned Teachers
}

\icmlsetsymbol{equal}{*}

\begin{icmlauthorlist}
\icmlauthor{Junyuan Hong}{equal,msu}
\icmlauthor{Yi Zeng}{equal,vt}
\icmlauthor{Shuyang Yu}{equal,msu}
\icmlauthor{Lingjuan Lyu}{sony}
\icmlauthor{Ruoxi Jia}{vt}
\icmlauthor{Jiayu Zhou}{msu}

\end{icmlauthorlist}

\icmlaffiliation{msu}{Michigan State University, Michigan, USA}
\icmlaffiliation{vt}{Virginia Tech, Virginia, USA}
\icmlaffiliation{sony}{Sony AI, Japan}

\icmlcorrespondingauthor{Lingjuan Lyu}{lingjuan.lv@sony.com}
\icmlcorrespondingauthor{Jiayu Zhou}{jiayuz@msu.edu}

\icmlkeywords{Machine Learning, ICML}

\vskip 0.3in
]

\printAffiliationsAndNotice{\icmlEqualContribution} %

\begin{abstract}
Data-free knowledge distillation (KD) helps transfer knowledge from a pre-trained model (known as the teacher model) to a smaller model (known as the student model) without access to the original training data used for training the teacher model. However, the security of the synthetic or out-of-distribution (OOD) data required in data-free KD is largely unknown and under-explored. In this work, we make the first effort to uncover the security risk of data-free KD w.r.t. untrusted pre-trained models. We then propose \uline{A}nti-\uline{B}ackdoor \uline{D}ata-Free KD (ABD), the first plug-in defensive method for data-free KD methods to mitigate the chance of potential backdoors being transferred. We empirically evaluate the effectiveness of our proposed ABD in diminishing transferred backdoor knowledge while maintaining compatible downstream performances as the vanilla KD. We envision this work as a milestone for alarming and mitigating the potential backdoors in data-free KD. 
Codes are released at 
\url{https://github.com/illidanlab/ABD}.
\end{abstract}

\section{Introduction}

In recent years, deep learning (DL) has witnessed tremendous success in solving real-world challenges \cite{yu2022coca,yamada2020luke,zhang2020pushing} by training huge models on giant data
\cite{dosovitskiy2020image,tolstikhin2021mlp,wang2022internimage}. Yet the performance-favored large model size has hindered their deployment to resource-limited~\cite{beyer2022knowledge} and communication-limited~\cite{tan2022towards} systems that meanwhile require responsive inferences, e.g., on tiny sensors, and frequent sharing of model parameters, e.g., federated learning~\cite{konevcny2016federated,zhu2021data}. 

To tailor the highly performant large models for the budget-constrained devices, knowledge distillation (KD) \cite{hinton2015distilling} and more recently data-free KD \cite{chawla2021data,ye2020data,fang2022up}, has emerged as a fundamental tool in the DL community. 
Data-free KD, in particular, can transfer knowledge from a pre-trained large model (known as the \emph{teacher model}) to a smaller model (known as the \emph{student model}) without access to the original training data of the teacher model. The non-requirement of training data generalizes KD to broad real-world scenarios, where data access is restricted for privacy and security concerns. 
For instance, many countries have strict laws on accessing facial images \cite{parkhi2015deep}, financial records \cite{shah2022flue}, and medical information \cite{antonelli2022medical}.
Recently, data-free KD also empowers federated learning on heterogeneous clients~\cite{zhu2021data,seo2022federated} and on low-bandwidth communication networks~\cite{zhu2022resilient,zhang2022fine,zhang2022dense}.

Despite the benefits of data-free KD and the vital role it has been playing, a major security concern has been overlooked in its development and implementation: 
\emph{Can a student trust the knowledge transferred from an untrusted teacher?}
The untrustworthiness comes from the non-trivial chance that pre-trained models could be retrieved from non-sanitized or unverifiable sources, for example, third-party model vendors~\cite{liu2022poisonedencoder} or malicious clients in federated learning~\cite{bagdasaryan2020backdoor}. 
One significant risk is from the \emph{backdoor} pre-implanted into a teacher model~\cite{jia2022badencoder}, which alters model behaviors drastically in the presence of predesigned triggers but remains silent on clean samples.
As traditional attacks typically require to poison training data \cite{gu2017badnets,souri2021sleeper,barni2019new,zeng2022narcissus}, it remains unclear if student models distilled from a poisoned teacher will suffer from the same threat without using the poisoned data.

\begin{figure}[!t]
    \centering
    \includegraphics[width=0.85\columnwidth]{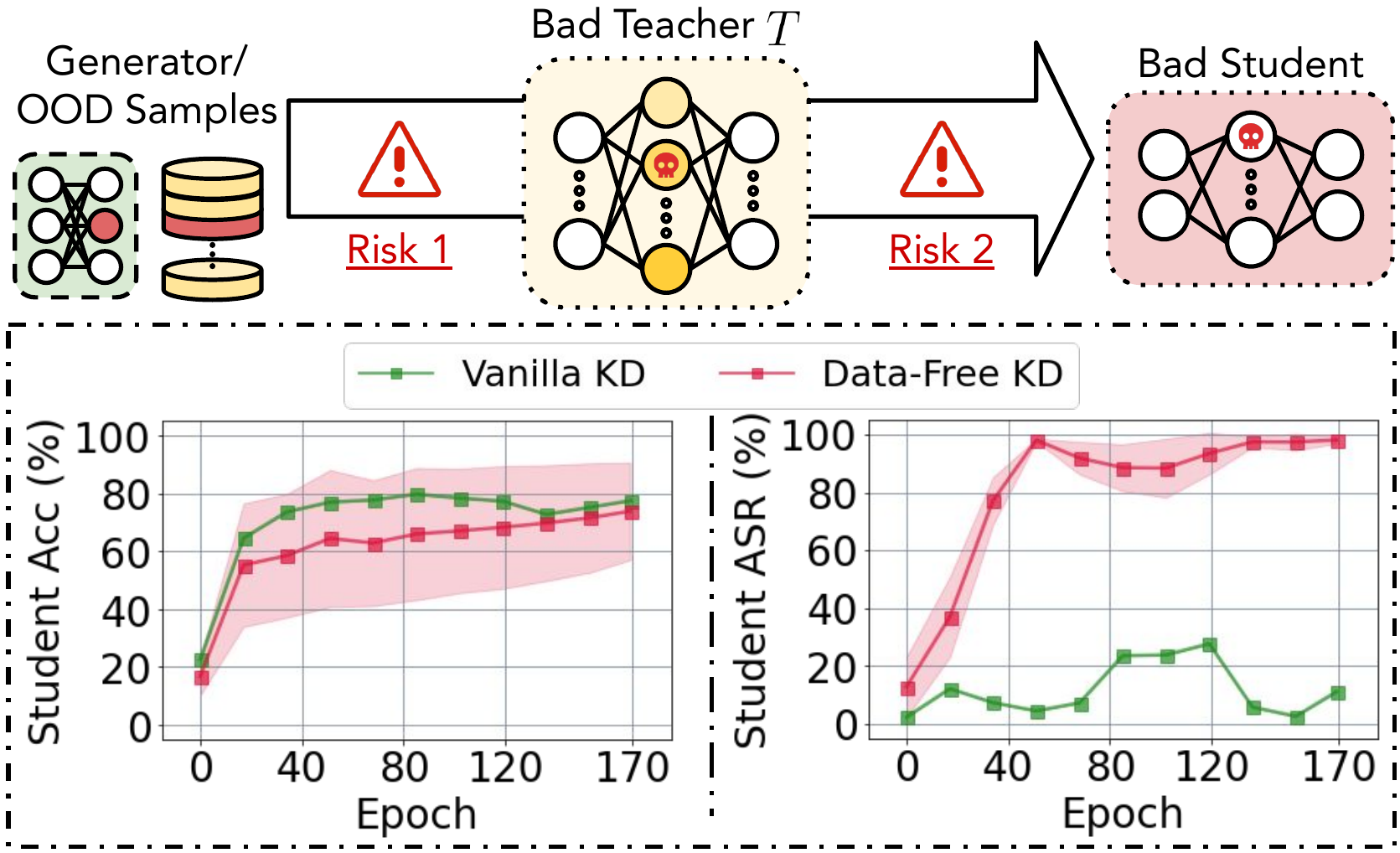}
    \caption{Data-free KD may transfer the backdoor knowledge (high Attack Success Rate, or ASR, from poisoned teachers to the student). The experiment is conducted on CIFAR-10 with a Trojan WM-poisoned teacher model~\cite{liu2017trojaning}. Vanilla KD denotes the KD with 10,000 clean in-distribution samples. The data-free KD is averaged with three methods that are based on synthetic or OOD data. The shadowed region is the standard deviation. We depict the student model's performance on clean samples (Student Acc) and samples patched with the backdoor trigger (Student ASR).
    }
    \label{fig:example_data_free}
    \vspace{-1.5em}
\end{figure}

In this paper, we take the first leap to uncover the \emph{data-free backdoor transfer} from a poisoned teacher to a student through comprehensive experiments on 10 backdoor attacks. %
We evaluated one vanilla KD using clean training data~\cite{hinton2015distilling} and three training-data-free KD method which use synthetic data (ZSKT~\cite{micaelli2019zero} \& CMI~\cite{fang2021contrastive}) or out-of-distribution (OOD) data as surrogate distillation data~\cite{asano2021extrapolating}.
To highlight the risks, we showcase the result of distilling a poisoned pre-trained WideResNet-16-2 \cite{zagoruyko2016wide} as the teacher on CIFAR-10 in \cref{fig:example_data_free}.
Our main observations in \cref{sec:risk} are summarized as follows and essentially imply two identified risks in data-free KD:
(1) Vanilla KD does not transfer backdoors by using clean in-distribution data, while all three training-data-free distillations suffer from backdoor transfer by 3 to 8 types of triggers out of 10 with a more than 90\% attack success rate.
Contradicting the two results indicates the \emph{poisonous nature of the surrogate distillation data} in data-free KD;
(2) The successful attack on distillation using trigger-free out-of-distribution (OOD) data demonstrates that triggers are not essential for backdoor injection, but the \emph{poisoned teacher supervision} is.

Upon observing aforementioned two identified risks, we propose a plug-in defensive method, \uline{A}nti-\uline{B}ackdoor \uline{D}ata-Free KD (ABD), that works with general data-free KD frameworks. ABD aims to suppress and remove any backdoor knowledge being transferred to the student, thus mitigating the impact of backdoors. The high-level idea of ABD is two-fold:
\textbf{Shuffling Vaccine (SV)} during distillation:~suppress samples containing potential backdoor knowledge being fed to the teacher (mitigating backdoor information participates in the KD); Student
\textbf{Self-Retrospection (SR)} after distillation:~ synthesize potential learned backdoor knowledge and unlearns them at later training epochs (the backstop to unlearn acquired malicious knowledge).
We believe ABD is a significant step towards making data-free KD secure and the downstream student trustworthy. The main contributions of this paper are as follows:
\vspace{-0.5em}
\begin{packeditemize}
    \item To the best of our knowledge, we are the first to uncover the security risk of data-free KD regarding untrusted pre-trained models on 10 backdoor types and 4 diverse distillation methods.
    \item We identify two potential causes for the backdoor infiltrating from the teacher to the student via data-free KD, that may inspire defense methods.
    \item To mitigate the data-free backdoor transfer, we propose ABD, the first plug-in defensive method for data-free KD methods. %
    \item To evaluate the effectiveness of the ABD, we conduct extensive experiments on 2 benchmark datasets and 10 different attacks to show ABD's efficacy in diminishing the transfer of malicious knowledge. 
\end{packeditemize}

\vspace{-1em}
\section{Background}

In this section, we introduce the preliminaries on backdoor attacks and data-free KD, and then we define the threat model considered in the paper.

\textbf{Backdoor attacks in pre-trained models.}
Backdoor attacks are an emerging security threat to DL systems when untrusted data/models/clients participate in the training process \cite{li2020backdoor}. 
Backdoor attacks
have developed from using sample-independent visible triggers \cite{gu2017badnets,chen2017targeted,liu2017trojaning} to more stealthy and powerful attacks with sample-specific \cite{li2021invisible} or visually imperceptible triggers \cite{li2020invisible,nguyen2021wanet,zeng2021rethinking,wang2022invisible}.
More advanced
attacks with clean labels ensure the manipulated features are semantically consistent with corresponding labels to better evade manual inspections \cite{turner2019label,souri2021sleeper,zeng2022narcissus}. 
The above backdoor attacks can be easily deployed to obtain a poisoned model, $T^*$, by minimizing:
\begin{equation}
\label{eqn:backdoor}
    \Ebb_{(\vx,y) \sim D} 
    \left[\underbrace{L(T(\vx),y)}_{\text{clean task}} + \underbrace{L(T(\vx+\delta),t)}_{\text{backdoor task}}
    \right],
\end{equation}
where $L$ is the cross-entropy loss; the clean task denotes the model performance on samples drawn from the clean distribution, $D$, without triggers (correctly classifying $x$ as label $y$); and the backdoor task denotes the malicious behavior of the model on observing samples patched with a trigger $\delta$ (classifying $x+\delta$ as the target label $t$). 
In this paper, we consider a case where the teacher model in KD is potentially inserted with a backdoor, and we focus on analyzing and resolving the associated security risks.

\begin{figure*} [!t]
\centering
  \includegraphics[width=\textwidth]{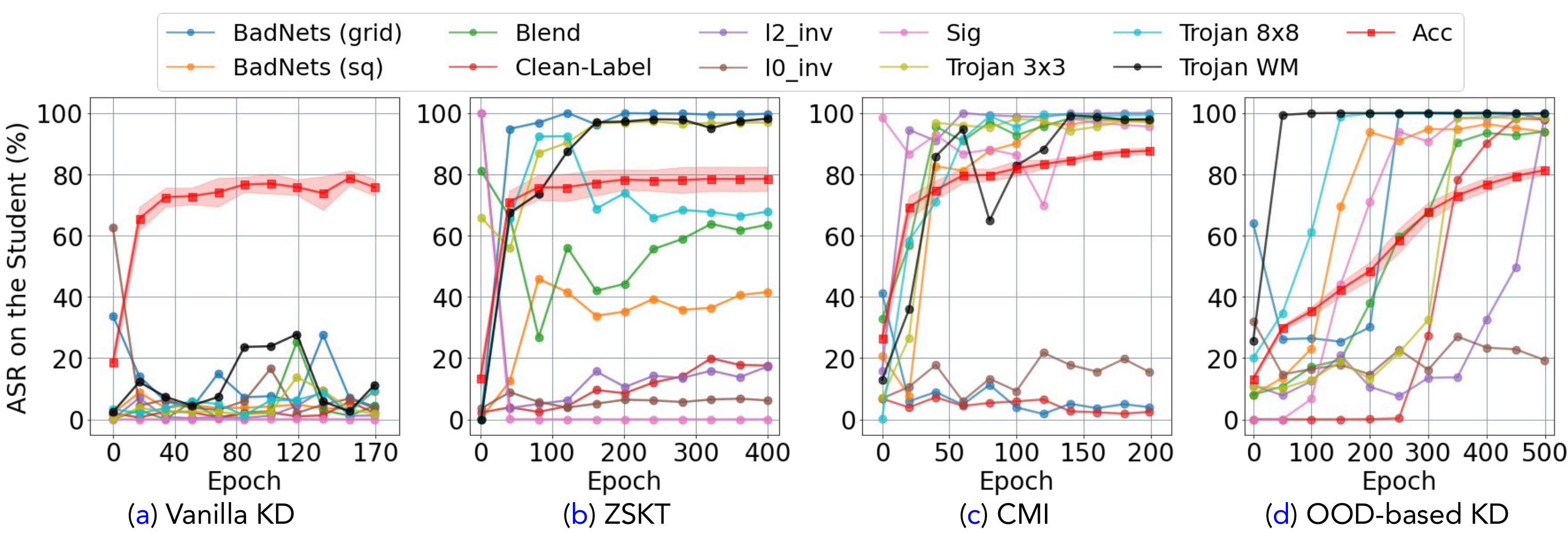}
  \captionof{figure}{
  Backdoor Attack Success Rates (ASRs) of the distilled student model using the vanilla KD with clean in-distribution samples (\textcolor{blue}{a}) and data-free KD using synthetic (\textcolor{blue}{b, c}) or OOD (\textcolor{blue}{d}) samples. The clean accuracy (Acc) of each figure is plotted with standard deviations among different attack-poisoned CIFAR-10.
  We run each KD method with different but sufficient training epochs to ensure convergence.
  Existing data-free KD methods may lead to the transfer of backdoor knowledge when poisoned teachers' participation.
  \label{fig:example_attacks}}
\end{figure*}

\noindent\textbf{Data-free knowledge distillation (KD).}
Without ambiguity, KD in this work refers to offline response-based KD \cite{hinton2015distilling}.
Given a teacher model, $T(\cdot)$, and some in-distribution samples (of the same distribution as the training data for the teacher), $\vx \sim D$, a typical optimization goal of response-based KD is to obtain student model parameters, $\theta$, that minimizes the Kullback-Leibler divergence loss, $D_{KL}(\cdot)$, between the output softmax-logits of the teacher and the student, $S(\cdot|\theta)$:
\begin{equation}
\label{eqn:KD}
\theta = \argmin\nolimits_{\theta} \Ebb_{\vx \sim D} \left[ D_{KL}\left(T\left(\vx \right) \| S\left(\vx|\theta\right)\right) \right].
\end{equation}
With data unavailability becoming more common in real-world DL settings due to privacy, legality, security, and confidentiality concerns, the development or implementation of KD has thus shifted to data-free settings.
The key difference between data-free KD and vanilla KD is that the samples used for KD are synthetic \cite{chen2019data,micaelli2019zero} or sampled from out-of-distribution (OOD) domains \cite{asano2021extrapolating}. 
Promising implementations of data-free KD have also been demonstrated in advanced federated learning frameworks \cite{tang2022virtual,zhu2021data,zhang2022fine}.
For generation-based methods, the dataset $D$ in \cref{eqn:KD} is replaced by a trainable data generator or a set of trainable images.
Generally, the dataset or generator can be parameterized as $P$ and trained by maximizing the disagreement between the teacher and student models:
\begin{align*}
    \max\nolimits_{P} \Ebb_{\vx \sim P} \left[ D_{KL}\left(T\left(\vx \right) \| S\left(\vx\right)\right) \right].
\end{align*}
Here, representative implementations include the first adversarial data-free distillation, Zero-Shot Knowledge Transfer (ZSKT) \cite{micaelli2019zero}, the state-of-the-art data-free KD methods, CMI \cite{fang2021contrastive}.
For OOD-based methods, we utilize the single-image extrapolation~\cite{asano2021extrapolating}, which extracts patches from a single image as training data for \cref{eqn:KD}.
For simplicity, we also denote the set of data as non-trainable $P$.

Even though techniques for data-free settings enable KD to be generalized more flexibly to data-constrained environments, no existing work has taken a closer look at the potential security risk of doing so. 
Given the fast development and emerging implementations of data-free KD for security concerning tasks, it is crucial to understand this security risk and study the countermeasures.

\textbf{Threat model: Knowledge of attacker and defender.}
For the purpose of risk evaluation and defense, we consider a standard security threat model where an un-trustworthy party participates in the teacher-training process. The \textbf{attacker} performs attacks \emph{only} by publicly releasing well-trained models inserted with backdoors or directly transmitting the poisoned model to the user (e.g., in federated learning). A user wishes to deploy the model's knowledge for further use but may require a different model structure due to size/memory constraints~\cite{wu2022communication}, %
or client heterogeneity \cite{zhu2021data}. 
For the \textbf{defender}, the original training data used by the attacker is unavailable
for knowledge transfer, and the goal is to develop a practical countermeasure to diminish the chance of transferring backdoor knowledge in data-free KD without additional knowledge requirements, i.e., the defender only has access to the teacher model.
Because of the data-free assumption, existing defenses requiring a clean dataset, for instance, \cite{li2021neural,zeng2021adversarial,wang2022trap} are typically excluded in our scenarios.

\section{Data-Free Can Steam Security Risks
}
\label{sec:risk}

\textbf{General Threats on Data-free Distillation.}
For an empirical evaluation of the existing data-free KD methods,
We consider 10 different backdoor attacks, BadNets with grid (grid) \cite{gu2017badnets}, BadNets (sq), Blend \cite{chen2017targeted}, Clean-label \cite{turner2019label}, l2-invisible (l2\_inv) \cite{li2020invisible}, l0-invisible (l0\_inv), Sig \cite{barni2019new}, 
Trojan Square $3\times 3$ (Trojan $3\times 3$), Trojan Square $8\times 8$ (Trojan $8\times 8$), and Trojan watermark (Trojan WM)~\cite{liu2017trojaning}. These attacks are then deployed to train 10 poisoned teacher models (WideResNet-16-2 \cite{zagoruyko2016wide}) with attack settings referred to in their original papers. The poisoned teacher models' performances and attack visualizations are provided in Figure \ref{fig:all_attack_cifar}. We further deploy different KD methods on these well-trained teacher models and obtain the respective student models for evaluation.
The results of these data-free KD methods are then compared to the vanilla KD, which uses 10,000 clean, in-distribution CIFAR-10 samples. We depict the attack success rate (ASR) of these attacks on the distilled student models in Figure \ref{fig:example_attacks}. The clean accuracy (Acc) on benign samples is similar when each method is run till converge (vanilla KD takes 170 epochs to converge. It takes 400, 200, and 500 epochs for ZSKT, CMI, and OOD-based methods to converge, respectively). We combined the Acc of all the models distilled by the respective KD method into a single red line with standard deviation.

From Figure \ref{fig:example_attacks}, we find that all the evaluated data-free KD approaches have transferred some of the attack's malicious knowledge from the poisoned teachers to the student. 
Based on the difference in the data used for knowledge distillation, we now highlight two potential risks that may lead to the transfer of backdoor knowledge. Additional results on other dataset-setting are presented in Appendix \ref{sec:app:risk}.

\noindent
\textbf{Potential Risk in Bad Synthetic Input Supply.}
Noting the Attack Success Rate (ASR) result on the students with vanilla KD using clean in-distribution samples is utterly different from the data-free settings' results. The key difference between the vanilla KD and data-free KD is the data supply, i.e., the input taken in by the teacher model. 
We hereby highlight a potential risk associated with the input supplied to the teacher in data-free KD. 
In particular, we find the poisoned teacher's participation in the synthetic data generation may lead to the generation of poisoned samples. 
We can assume the student starts without backdoor knowledge, i.e., $S(\vx + \delta_t) = S(\vx)$\footnote{We omit $\theta$ when $S(\cdot)$ is deployed for evaluation for simplicity} for any $\vx$.
We may simplify the data generation by maximizing the error by the student:
$
    \vx_p = \argmax_{\vx} D_{K L}\left(T\left(\vx \right) \| S\left(\vx\right)\right).
$
We may reformulate $\vx$ as $\vx = \vx_0 + \delta$ and assume $\vx_0 \in \{ \vx | T(\vx)\neq t\}$.
We assume $\delta \in C_{<\epsilon}$, which is a potential backdoor within a bounded constraint, e.g., $\norm{\delta} \le \epsilon$.
Note that though there is no constrained optimization in practice, the small learning rate and uncontrolled optimization may converge into a pitfall.
Thus, equivalently,
$
    \delta_p = \argmax_{\delta\in C_{<\epsilon}} D_{K L}\left(T\left(\vx_0 +\delta \right) \| S\left(\vx_0 + \delta\right)\right).
$
Note that 
\begin{align*}
    &\quad D_{K L} \left(T\left(\vx_0 +\delta_t \right) \| S\left(\vx_0 + \delta_t \right)\right) \\
    &= D_{K L}\left(T\left(\vx_0 +\delta_t \right) \| S\left(\vx_0\right)\right) \ge D_{K L}\left(T\left(\vx_0 \right) \| S\left(\vx_0\right)\right)
\end{align*}
Therefore, there is a chance to generate $\delta_p$ from the above maximization, i.e.,
$
    P[\norm{ \delta_t - \delta_p } \le \epsilon] > 0.
$
In other words, there is a potential risk associated with the input supplied to the teacher for distillation in data-free KD.

\noindent
\textbf{Potential Risk in Bad Supervision.}
On the other hand, the process of sampling data from OOD does not have the poisoned teacher's participation. However, we still find attacks that can infiltrate the teacher to the student via OOD-based KD. We hereby highlight another potential risk associated with the output logits of the teacher. That is, the returned soft labels may contain backdoor knowledge and thus lead to bad students.

\vspace{-1em}
\section{Anti-Backdoor Data-Free KD}

\begin{figure}[!t]
  \centering
  \includegraphics[width=\linewidth]{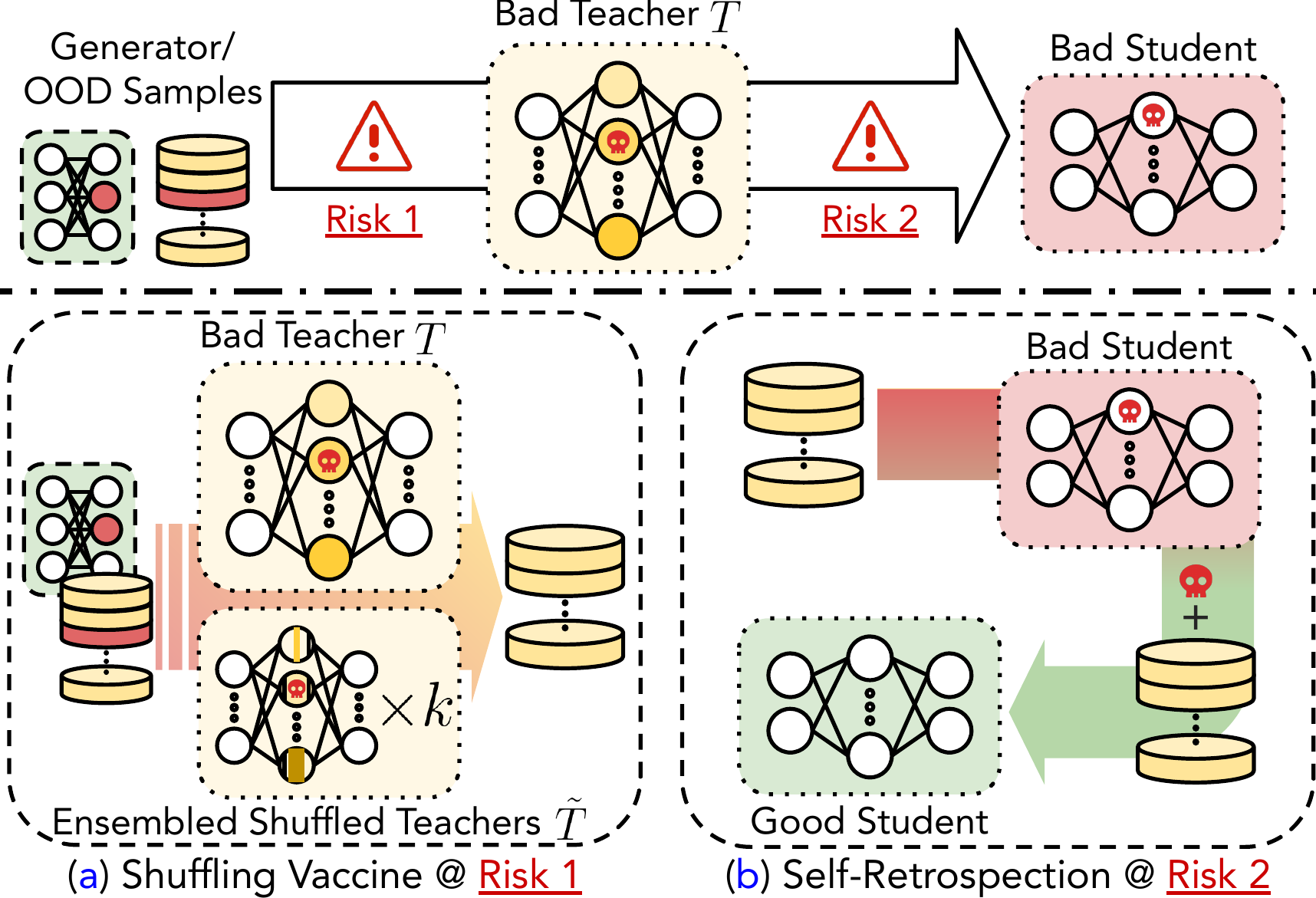}
  \caption{Risks of standard data-free KD with generator/OOD samples and the proposed ABD. \uline{Upper part}: the two identified risks in data-free KD. \uline{Lower parts}: the proposed ABD. (\textcolor{blue}{\textbf{a}}) \textbf{Shuffling Vaccine} diminish the chance of bad input supplied for distillation; (\textcolor{blue}{\textbf{b}}) student \textbf{Self-Retrospection} at a later stage of training to confront the potential learned backdoor from the teacher.
  }
  \label{fig:abd_flowchart}
\end{figure}

On observing the significant risks, we propose a plug-in anti-backdoor fixture for securing the existing data-free distillation method as formulated in \cref{eqn:KD}.
Our method is composed of two sequential strategies aimed to mitigate the two potential risks discussed in \cref{sec:risk}: \textbf{Shuffling Vaccine (SV)} before distillation optimization to diminish the chance of potential backdoored samples' participation and \textbf{Self-Retrospection (SR)} of the student at a later training stage of the student model to confront the bad supervision. An overview of our method and the relation to the two identified risks is illustrated in \cref{fig:abd_flowchart}.

\subsection{Shuffling Vaccine (SV)}

Our method is inspired by the recent advance on backdoor model suspection~\cite{cai2022randomized}.
Previous work has revealed \cite{tran2018spectral,hayase2021spectre} the sparse nature of backdoor activations.
Most images will activate different feature channels in deep layers of a network.
In contrast, backdoor triggers will singly but significantly light a few channels such that other semantic features will be weakened layer by layer.
As the backdoor activation is sparse, Cai \etal~proposed \emph{Channel Shuffling}, which amplified the nature by shuffling channels to suspect if a model is ever poisoned.
The intuition is that the backdoor only relies on a few channels and shuffling may not destroy the connection.
Instead, the prediction path for clean images will be ruined since a high ratio of semantic features will be compromised.

\begin{figure}[!t]
    \centering
    \includegraphics[width=0.48\columnwidth]{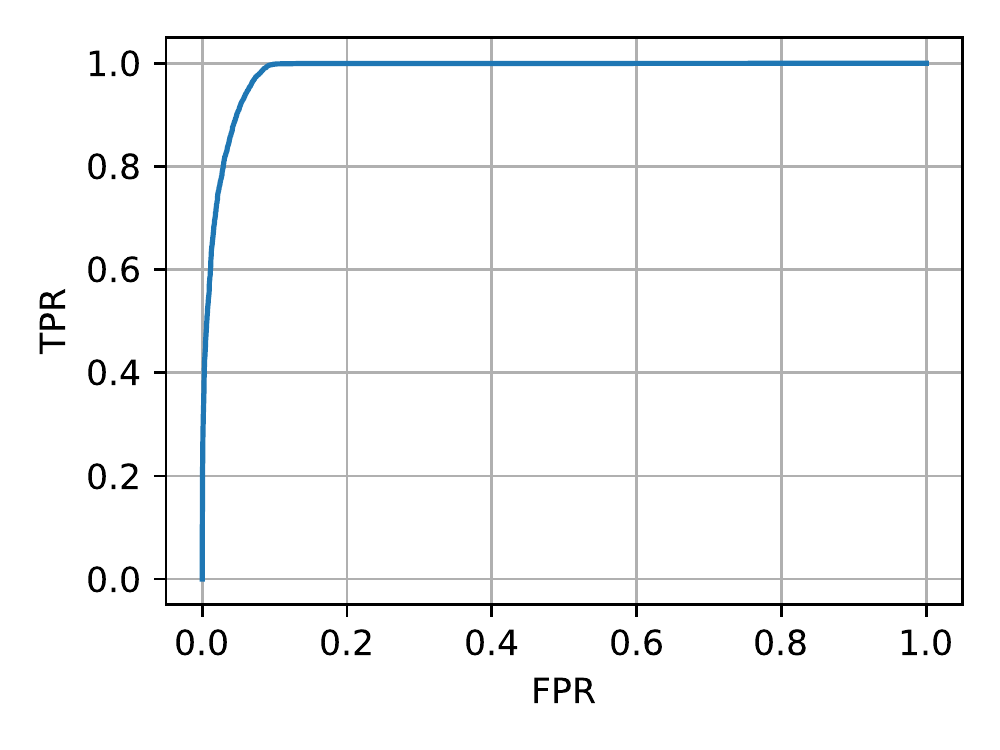}
    \includegraphics[width=0.48\columnwidth]{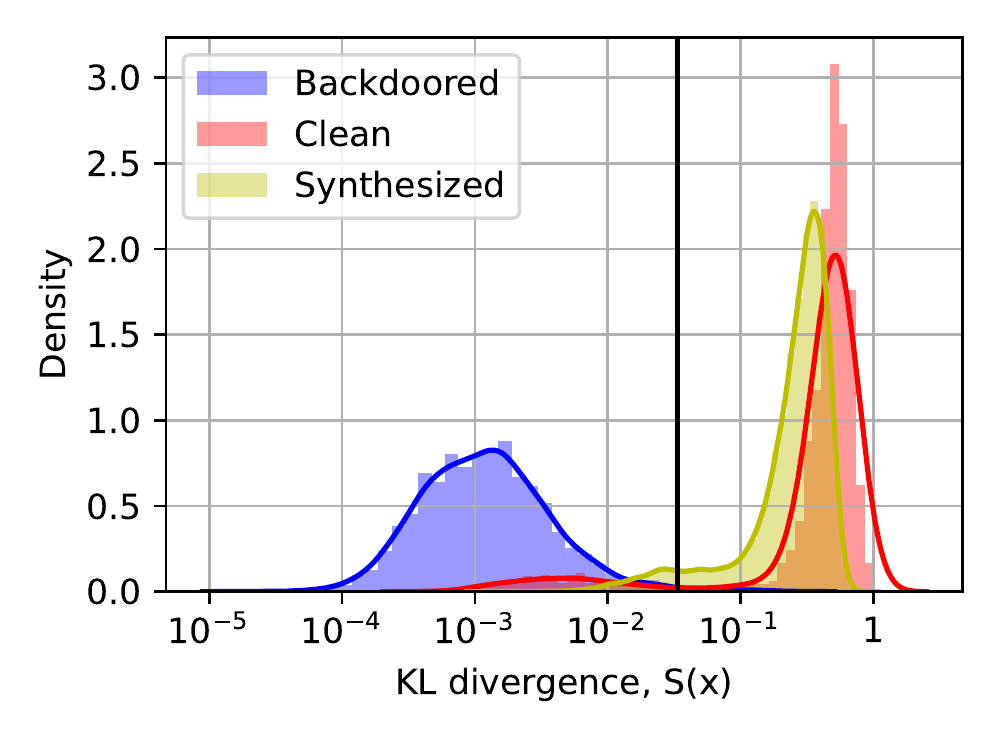}
    \vspace{-1em}
    \caption{(a) ROC curve of $\mathcal{S}(x)$ colored by clean or backdoored samples. The corresponding AUC is 0.984. (b) Comparing $\mathcal{S}(x)$ where the black vertical line represents the $3\sigma$ boundary of the backdoored samples. A portion of the synthetic images falls into the danger zone.
    }
    \label{fig:poison}
\end{figure}

In this work, we novelly repurpose the \emph{Channel Shuffling} to detect suspicious samples that may rely on some shortcuts in the networks.
We hypothesize that if a sample can be stably predicted as one class under Channel Shuffling, then the sample is prone to be poisonous.
Formally, we derive a shuffled model $\tilde T$ by shuffling the last few layers of $T$ and then define a score metric as: 
\begin{align*}
    \mathcal{S}(x;\tilde T) = \log D_{KL}(\tilde T(x) \| T(x)),
\end{align*}
where a smaller value indicates a higher risk of poisoning.
In \cref{fig:poison}, we show that triggered samples have much lower $\mathcal{S}(x)$ than clean samples, showing that the metric is effective for detecting poison samples;
Therefore, we apply the metric to suppress the generation or usage of suspicious samples, such that we can mitigate backdoor transfer in data-free distillation.

\textbf{(1)~Suppresing backdoor generation.}
For methods like ZSKT and CMI,  distillation is built upon a synthetic dataset by contrasting the teacher and student models.
Given a teacher model $T$ and a shuffled model $\tilde T$, we define a new regularization term, $R(\vx; \tilde T, T)$ on synthetic samples $\vx$:
\begin{gather*}
    \max_{P} \Ebb_{\vx \sim P} \left[ D_{KL}\left(T\left(\vx \right) \| S\left(\vx\right)\right) + \alpha R(\vx; \tilde T, T) \right] , \\
    R(\vx; \tilde T, T):= \phi(T\left(\vx \right) \| \tilde T\left(\vx\right)) D_{KL}\left(T\left(\vx \right) \| \tilde T\left(\vx\right)\right),
\end{gather*}
where $\phi(T\left(\vx \right) \| \tilde T\left(\vx\right))$ will yield 1 if the predicted labels of $T(\vx)$ and $\tilde T(\vx)$ are the same.
Considering the randomness of shuffling, we use an ensemble of three shuffled teachers as $\tilde T$.

\textbf{(2)~Suppressing suspicious distillation.}
For OOD distillation, there is no way to control the sample generation and thus calls for a different defense method.
One straightforward way is to drop the suspicious samples, which however may reduce the data size and result in overfitting.
Instead, we introduce a soft constraint on the distillation to better trade-off model utility and security.
\begin{align*}
    \min_{\theta} \Ebb_{\vx \sim P} \left[ (1-\phi + \tfrac{1}{\alpha} \phi)D_{KL}\left(T\left(\vx \right) \| S\left(\vx\right)\right) \right],
\end{align*}
where $\phi$ is the output of $\phi(T\left(\vx \right) \| \tilde T\left(\vx\right))$.
If $\phi$ is activated, then the sample loss will be shrunk by $\alpha$.

\subsection{Self-Retrospection (SR)} 
To mitigate the potential risk associated with bad supervision, we propose a post-hoc treatment to use the student's knowledge to confront potential backdoor knowledge that has been learned from the teacher.
More specifically, we use the student's own knowledge to synthesize potential backdoor knowledge being learned and confront the model update from the teacher's supervision with the following Self-Retrospection (SR) task:
\begin{equation*}
    \theta^* = \argmin_\theta \max_{\delta \in C_{<\epsilon}}\frac{1}{ n }\sum_{i=1}^{n}
    D_{KL}\left(S\left(\vx |\theta\right) \| S\left(\vx+\delta|\theta\right)\right),
\end{equation*}
noting that $\delta$ in the outer loop is a function that depends on $\theta$ in the inner loop, i.e., $\delta(\theta)$.
The intuition of the formulation is to synthesize a universal noise that will result in most of the samples' output logits greatly changed compared to the output of these samples without the noise. The amount of change is then depicted by the KL divergence, as a larger value of KL divergence depicts a stronger variation between the outputs with or without the noise being patched.

To resolve the proposed bi-level optimization, inspired by \citet{zeng2021adversarial,rajeswaran2019meta}, we approximate $\nabla \delta(\theta)$ with a suboptimal solution of $\delta^*$. In particular, one can approximate $\nabla \delta(\theta)$ with an iterative solver of limited rounds (e.g., conjugated gradient algorithm \cite{rajeswaran2019meta}, or fixed-point algorithm \cite{grazzi2020iteration}) along with the reverse mode of automatic differentiation \cite{griewank2008evaluating}. With a successful estimation of  $\nabla \delta(\theta)$, we then can plug it into the process of computing the complete hypergradient for student SR:
\begin{align*}
    \nabla \psi(\theta)
    = 
\nabla_{2}D_{KL}(\delta(\theta),\theta)
+
\left(\nabla \delta(\theta)\right)^{\top} 
\nabla_{1}{D_{KL}(\delta(\theta), \theta)}
\end{align*}
where we simplified $D_{KL}\left(S\left(\vx |\theta\right) \| S\left(\vx+\delta|\theta\right)\right)$ as a function to $\delta$ and $\theta$, where $\delta$ is a variable dependent on $\theta$, i.e., $D_{KL}(\delta(\theta), \theta)$, and
$\nabla_1(\cdot)$ or $\nabla_2(\cdot)$ denotes the partial derivatives w.r.t. the first variable or the second variable respectively. We summarize the whole process of one round of student SR in \cref{algo:unlearning}, where our synthesized student SR hypergradeint is used to confront the original gradient acquired from $D_{KL}\left(T\left(\vx\right) \| S\left(\vx|\theta\right)\right)$ for student update thus mitigates the potential risk of bad supervision.

\SetKwInput{KwParam}{Parameters}
\begin{algorithm}
\small
    \caption{One Round of KD with Self-Retrospection}
    \label{algo:unlearning}
    \SetNoFillComment
    \KwIn{
    $T(\cdot)$ (Teacher model);
    \\ \quad \  \qquad  
    $S(\cdot;\theta)$ (Student model with parameters $\theta$);
    }
    \KwParam{
    $n_{\delta}$ (Number of steps);
    \\  \quad \quad \quad  \quad \quad \quad 
    $\eta,\gamma > 0$ (Step size);
    }
    \BlankLine

                $\mathcal{L}_{S}\leftarrow D_{KL}\left(T\left(\vx\right) \| S\left(\vx|\theta\right)\right)$\;
                
                $\delta\sim\mathcal{N}(0,\sigma^2\mathbf{I}^d)$\;
                
                \For{$1,2,\ldots,n_{\delta}$}{
                    $\mathcal{L_{\delta}}\leftarrow -D_{KL}\left(S\left(\vx |\theta\right) \| S\left(\vx+\delta|\theta\right)\right)$\;
                    
                    $\delta\leftarrow\delta-\gamma\frac{\partial\mathcal{L}_{\delta}}{\partial\delta}$\;
                     }
    
                Estimate $\nabla \delta^{\top}$ by assuming $\delta$ is suboptimal with iterative solver\;

                Compute $\nabla \tilde{\psi}(\theta)$ with $\nabla \delta^{\top}$ pluged in\;
    
                $\theta\leftarrow \theta - \eta\left( \frac{\partial\mathcal{L}_S}{\partial\theta} + \nabla \tilde{\psi}(\theta) \right)$\;

\end{algorithm}

\subsection{Overall Pipeline}%

\textbf{Vaccine verification and search.}
Due to the random nature of shuffling and backdoor mechanisms, there is a chance that the Shuffling Vaccine is not able to detect triggers.
Therefore, we verify the functionality of Shuffling Vaccines before using them.
The challenge of verification is lacking known clean and poisoned samples.
For this purpose, we first run data-free KD to cache some surrogate data as set $D_{s}$ and check if the $\cS(x)$ distribution has a large tail.
The intuition has been illustrated in \cref{fig:poison}.
According to the three-sigma rule of thumb, a normal distribution should have 0.3\% samples of values smaller than $\mu -3\sigma$ where $\mu$ and $\sigma$ are mean and standard deviations, respectively.
Thus, we check the existence of the tail by computing the tail ratio, defined as
\begin{align*}
    \tau(\tilde T; D_s) := \tfrac{| \{x \in D_s |\cS(x;\tilde T) < \mu - 3\sigma\} | }{|D_s|},
\end{align*}
where $\mu$ and $\sigma$ denote the mean and standard deviation of $\{\cS(x;\tilde T) | x\in D_s\}$.
We threshold $\tau(\tilde T; D_{s})$ by $0.02$ to choose a shuffle model with a large tail.
If $\tilde T$ does not satisfy the condition, we will repeat shuffling for 8 times until giving up. The setting of the threshold is further ablated in Appendix \ref{sec:app:ablation}.

We summarize our whole pipeline in the \cref{algo:concentration}.
We first use Shuffling Vaccines if a proper vaccine can be found.
If a vaccine is not found, we will do normal data-free KD and use the student SR as a post-hoc treatment at a later learning stage when the student is well-trained till converged.
If a vaccine is found, we may ask the user to determine if a sacrifice of clean accuracy is worth it for better security and activates the student SR on demand.
In our setting, we activate SR if the clean accuracy drops lower than 5\% using SV.

\noindent
\textbf{Time complexity analysis.}
SV is utilized to obtain an ensemble of effective shuffled models, and the forward pass of these models is used to suppress backdoor information. Compared to vanilla data-free KD for each epoch that includes SV, we introduce an additional $\tilde{O}(n \cdot \tilde{O}(\theta_{T}))$ time complexity, where $\tilde{O}(\theta_{T})$ represents the time complexity of using the teacher model, $\theta_{T}$, in a single forward pass on a batch of data. $n$ is the number of shuffle models used in the ensemble.
For SR, based on our Algorithm \ref{algo:unlearning} design (total $n_{\delta}$ rounds) and assuming the fixed-point algorithm \cite{grazzi2020iteration} as the iterative solver with $\vartheta$ iterations for computing $\nabla \tilde{\psi}(\theta)$, the time complexity is
 $\tilde{O}(n_{\delta} \cdot \vartheta \cdot \tilde{O}(\theta))$, where $ \tilde{O}(\theta)$ is the time complexity of training the student model, $\theta$, via backpropagation on a batch of data (similar to the forward pass of one epoch
 given the same quantity of samples \cite{zeng2021adversarial}). In practice, we adopted $\vartheta=5$, and for most one-target attack cases, $n_{\delta}=10$ is sufficient for algorithm convergence. Both techniques only introduce linear additional computational costs on the order of the size of the teacher or student model. In practice, we find in our experiment, the overall ZSKT+ABD empirical time cost with our settings on the WRNs to be only 1.03 times higher than the vanilla ZSKT, evaluated on CIFAR-10 with ZSKT and BadNets (grid) trigger.

\SetKwInput{KwParam}{Parameters}
\begin{algorithm}[!h]
\small
    \caption{Anti-Backdoor Data-Free KD (ABD)}
    \label{algo:concentration}
    \SetNoFillComment
    \KwIn{
    $T(\cdot)$ (Teacher model);
    \\ \quad \  \qquad  
    $S(\cdot;\theta)$ (Student model with parameters $\theta$);
    }
    \KwParam{
    $\lambda$ (Starting step for student SR);
    }
    \BlankLine
    
    Synthesize or obtain a set of OOD samples $D_s$\;
    
    Search for $\tilde{T}$ at most $8$ trials\; %
    
    \uIf{Found effective $\tilde T$}{
        \tcc{1. Early Prevention with SV}
    
        Data-free KD with SV till step $\lambda$\;
        
    }
    \Else{
    
    Data-free KD till step $\lambda$\;

    }
    \tcc{2. Later Treatment with SR}
    \If{Activates Student SR}{
    
    Data-free KD with student SR\;
    
    }
        
\end{algorithm}

\section{Experiment}

In this section, we evaluate how the proposed ABD can secure data-free KD against backdoor attacks under various data, model, and trigger configurations.

\textbf{Datasets and models.}
We use the same datasets, CIFAR-10 \cite{krizhevsky2009learning} and GTSR-B \cite{stallkamp2012man}, as \cite{zeng2021adversarial} to evaluate the backdoor defenses.
Following the setup of ZSKT~\cite{micaelli2019zero}, we use WideResNet~\cite{zagoruyko2016wide} for training 
10-way or 43-way classifiers on CIFAR-10 and GTSR-B, respectively.
We use WRN-16-2 to denote a 16-layer WideResNet with a width factor of $2$.

\begin{figure*}%
\centering
  \includegraphics[width=0.7\textwidth]{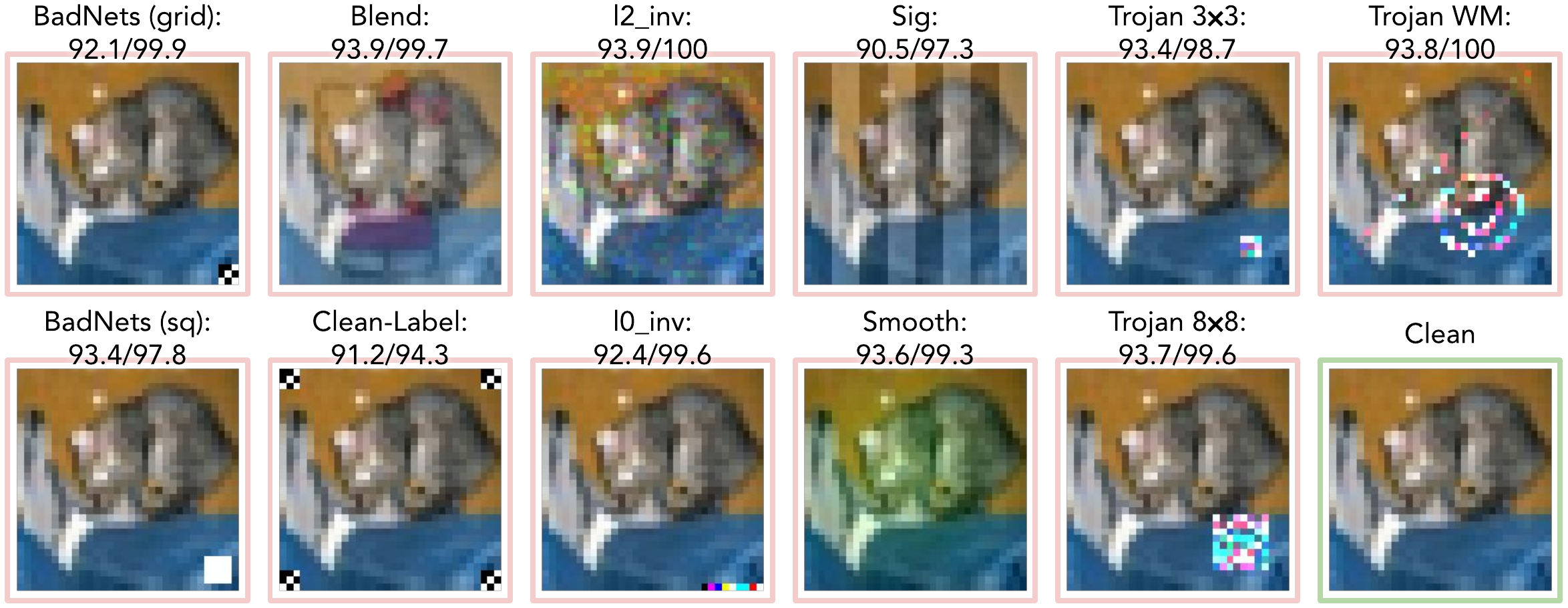}
  \captionof{figure}{
  Trigger visualization and teacher model performances on CIFAR-10. The performance (Acc/ASR) of the poisoned teacher using each backdoor attack is provided beneath each trigger's name. We envision the backdoored example for each attack on CIFAR-10.
  \label{fig:all_attack_cifar}}
\end{figure*}

\textbf{Backdoor attacks.} Prior to distillation, we pre-train a teacher model on a poisoned training dataset and use data-free distillation methods to train a student under the soft supervision of the pre-trained teacher model.
The poisoned pre-training dataset contains 10\% samples with backdoors injected by different attack manners.
For example, the BadNets attack~\cite{gu2017badnets} injects \underline{grid} or \underline{sq}uare patterns to the corner of an image, which are denoted as BadNets (grid) or (sq).
Examples of triggers are presented in \cref{fig:all_attack_cifar}.

\textbf{Evaluation metrics.} Following the common practice on backdoor defense, e.g., \cite{li2020neural,li2021anti,zeng2021adversarial}, we use attack success rate (ASR) and clean accuracy (Acc) as the measures evaluating distillation methods.
ASR is defined as the portion of backdoored test samples that can
successfully mislead the model to predict the target class specified by the attacker. 
Acc is %
the classification accuracy measured on a clean test set. 
A favored method should present a smaller ASR and meanwhile a larger Acc.

\textbf{Distillation methods.}
We use ZSKT~\cite{micaelli2019zero}, CMI~\cite{fang2021contrastive}, and OOD~\cite{asano2021extrapolating} as the baseline distillation methods.
We use 20\% clean data for vanilla knowledge distillation~\cite{hinton2015distilling}, denoted as Clean KD.
We follow previous work to use their published codes\footnote{\url{https://github.com/polo5/ZeroShotKnowledgeTransfer}}\footnote{\url{https://github.com/zju-vipa/CMI}}\footnote{\url{https://github.com/yukimasano/single-img-extrapolating}} and hyperparameters.
More details in \cref{sec:app:exp}.

\begin{table}[t]
    \centering
    \small
    \setlength\tabcolsep{4.5 pt}
    \scalebox{0.9}{
    \begin{tabular}{c|c|ccc}
      \toprule
     Trigger & Teacher & \multicolumn{3}{c}{Student Acc/ASR} \\
      & Acc/ASR & ZSKT  & ZSKT+ABD & Clean KD \\
      \midrule
      BadNets (grid) & 92.1/99.9 & 71.9/96.9 & 68.3/0.7 & 74.6/4.3 \\
      Trojan WM     & 93.8/100  & 82.7/93.9 & 78.2/22.5 & 77.5/11.1 \\
      Trojan 3x3   & 93.4/98.7 & 80.9/96.8 & 71.7/33.3 & 72.9/1.7 \\
      Blend        & 93.9/99.7 & 77.0/74.4 & 71.5/23.1 & 78.0/4.3  \\
      Trojan 8x8   & 93.7/99.6 & 80.5/57.2 & 72.6/17.8 & 75.2/9.3 \\
      BadNets (sq) & 93.4/97.8 & 80.8/37.8  & 77.9/1.9 (s) & 76.2/9.1 \\
      CL           & 91.2/94.3 & 76.8/17.5 & 67.4/10.2 & 69.4/2.1 \\
      Sig          & 90.5/97.3 & 77.9/0.0  & 72.2/0. (s) & 77.4/0. \\
      l2\_inv       & 93.9/100  & 82.0/0.3  & 70.7/1.9 (s) & 77.2/1.2 \\
      l0\_inv       & 92.4/99.6 & 72.8/8.3  & 69.4/0. (s) & 79.2/3.7 \\
      \bottomrule
    \end{tabular}}
    \caption{Evaluation of data-free distillation on more triggers on CIFAR-10 with WRN16-2 (Teacher) and WRN16-1 (student). (s) indicates Shuffling Vaccine is used instead of student SR.}
    \label{tbl:vary_trigger}
\end{table}

\begin{table*}[ht]
    \centering
    \small
    \begin{tabular}{c|ccc|c|ccc}
      \toprule
      Dataset & Teacher & Student & Teacher & Teacher & \multicolumn{3}{c}{Student Acc/ASR} \\
              & Arch (size) & Arch (size) & Trigger & Acc/ASR & ZSKT  & +ABD & Clean KD \\
      \midrule
      GTSR-B    & WRN16-2 (0.7MB) & WRN16-1 (0.2MB) & BadNets (grid) & 88.1/98.8 & 87.0/99.5 & 78.4/13.0 & 89.8/0.3 \\
      \midrule
      \multirow{4}{*}{CIFAR-10} & WRN16-2 (0.7MB) & WRN16-1 (0.2MB) & BadNets (grid) & 92.1/99.9 & 71.9/96.9 & 68.3/0.7 & 74.6/4.3  \\
              & WRN16-2 (0.7MB) & WRN16-1 (0.2MB) & Trojan WM & 93.8/100 & 82.7/93.9 & 78.2/22.5 & 77.5/11.1 \\ %
      & WRN40-2 (2.2MB) & WRN16-1 (0.2MB) & BadNets (grid) & 94.5/100 & 84.2/4.6 & 76.9/10.7 (s) & 72.0/4.7 \\ %
              & WRN16-2 (0.7MB) & WRN16-1 (0.2MB) & Trojan WM & 94.5/100 & 87.6/54.5 & 82.9/5.8 (s) & 71.2/5.3 \\ %
      \bottomrule
    \end{tabular}
    \caption{Evaluation of anti-backdoor data-free distillation on different datasets and different model architectures. `(s)' indicates Shuffling Vaccine is used instead of the student's Self-Retrospection.
    }
    \label{tbl:benchmark_zskt}
\end{table*}

\textbf{Defending multiple types of attacks.}
To evaluate the effectiveness of the proposed defending method, we construct a benchmark against different backdoors.
We use WRN16-2 as the teacher and WRN16-1 as the student to predict image classes in CIFAR-10.
All the teachers are trained and selected based on the best test accuracy on clean images.
The results are summarized in \cref{tbl:vary_trigger}.
In most triggers, our method effectively treats or protects the ZSKT from backdoor transfer successfully by reducing ASR lower than 30\%.
Since there is no free lunch for removing backdoors, reducing ASR also results in lower clean accuracy.
Especially, without clean data from the training set of teacher models, removing backdoors is even harder to maintain accuracy as compared to the clean KD.
This is because, without data from the same distribution, it is hard to distinguish which kinds of features are needed for benign tasks or backdoor tasks.
To effectively suppress the risks of backdoors, the degradation of clean accuracy is the essential cost.

\textbf{Data-free distillation has almost-free resilience to some backdoors.}
In \cref{tbl:vary_trigger}, we observe that many triggers are not strong enough to transfer without data.
The failure happens when the triggers are not localized to a small region but spatially spreading, e.g., Sig, CL, and l2\_inv.
Noticeably, the CL trigger relies on the adversarial samples to transfer, which fails with smoothed decision boundaries defined by the teachers' soft labels~\cite{yuan2020revisiting}.
Remarkably, the natural resilience is almost free for distillation, since the distillation does not significantly reduce accuracy compared to the strong transferred cases.

\textbf{Architectures and datasets in distillation.}
In \cref{tbl:benchmark_zskt}, we evaluate our method on defending ZSKT against BadNets (grid) and Trojan WM on two datasets and two teacher architectures.
There are several intriguing observations.
(1) Except for the BadNets from WRN40-2 teacher, all the triggers successfully transfer from the teacher to the student.
(2) We notice that the transfer is less effective when the teacher has deeper layers, comparing the WRN40-2 versus WRN16-2 teachers on CIFAR-10.
In such under-transfer cases, the anti-backdoor ZSKT even outperforms the clean KD.
This may imply that a deeper and over-parameterized model may be more robust in transferring clean knowledge than a few really-clean samples.
(3) In all cases, our method effectively defends ZSKT against the tested backdoor attacks.
Our method can maintain higher clean accuracy on CIFAR-10, which is composed of more complicated features than the traffic signs in GTSR-B.
This observation is surprisingly different from the one in central defense, e.g., in \cite{zeng2021adversarial} where CIFAR-10 is a harder dataset to defend.
The rationale behind the difference is that GTSR-B encodes simple features sparsely in the same network leaving more space for the trigger features.
Therefore, adversarial training by ZSKT will rely more on these triggers to transfer knowledge.
Instead, CIFAR-10 models can squeeze out these features to maintain good performance on clean images.

\begin{table}[ht]
    \centering
    \small
    \scalebox{0.9}{
    \begin{tabular}{c|cc|cc}
      \toprule
      Distillation & Teacher & Teacher & \multicolumn{2}{c}{Student Acc/ASR} \\
      Method & Trigger & Acc/ASR & Baseline &  +ABD \\
      \midrule
       \multirow{2}{*}{ZSKT} & Trojan WM & 93.8/100 & 82.7/93.9 & 78.2/22.5 \\
        & BadNets (grid) & 92.1/99.9 & 71.9/96.9 & 68.3/0.7  \\
       \midrule
       \multirow{2}{*}{CMI} & Trojan WM & 93.8/100 & 89.1/99.0 & 79.8/8.0 \\ %
        & BadNet (sq) & 93.8/100 & 88.3/95.9 & 83.2/6.0 \\ %
       \midrule
       \multirow{2}{*}{OOD} & Trojan WM & 93.8/100 & 82.3/100 & 62.3/21.8 \\ %
        & BadeNet (grid) & 92.1/99.9 & 79.8/99.6 & 78.2/14.5 \\ %
      \bottomrule
    \end{tabular}}
    \caption{ABD is effective in different data-free distillation methods on CIFAR-10 with WRN16-2 (Teacher) and WRN16-1 (student). 
    }
    \label{tbl:multi_distill}
\end{table}

\begin{figure}[!h]
    \centering
    \includegraphics[width=0.48\textwidth]{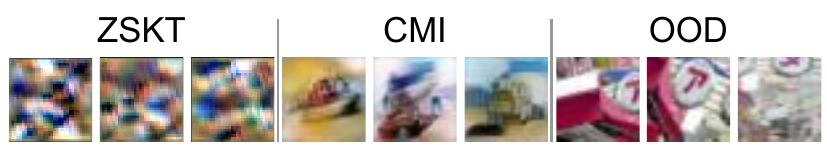}
    \vspace{-0.1in}
    \caption{Examples of ZSKT and CMI synthetic images. OOD images are patches of a single large image.}
    \label{fig:examples}
\end{figure}

\textbf{Protecting distillation with different surrogate data.}
As the data-free knowledge transfer heavily relies on the surrogate data for distillation, we investigate the resilience of different distillation methods and evaluate the backdoor defense on the failure cases.
In \cref{fig:example_attacks}, we have witnessed the backdoors can transfer successfully through ZSKT, CMI, and OOD.
In \cref{tbl:multi_distill}, we compare them on the selected transferred triggers.
With the same trigger of Trojan WM, all three distillation approaches have >90\% ASR.
Similar to ZSKT, CMI uses the adversarial loss to find the underfitted samples and therefore transfers knowledge by distilling these synthetic images.
Different from ZSKT, CMI introduces more objectives in data optimization to improve the quality of synthetic data, as visually compared in \cref{fig:examples}.
Unfortunately, improving visual quality cannot eliminate triggers even these triggers could be thought as visually-low-quality features.
Interestingly, our method can maintain higher clean accuracy on CMI compared to ZSKT and OOD.
Among the three distillations, CMI has traded the least amount of benign accuracy for lower ASR.
This implies that image quality is essential for benign accuracy though not robustness.

Compared to ZSKT, OOD is more vulnerable to the distributed trigger, Trojan WM, but is more robust against Badnet (grid), by using our defense.
This may be attributed to that the local BadNets trigger is less likely to be found in augmented real images than adversarially-generated ones by ZSKT.
Instead, Trojan WM has a pattern similar to random noise, that can be approximated by some random pattern in some image augmentations.

\textbf{Ablation study.}
A natural question for the proposed defense may be \emph{why we need student SR as a post-hoc treatment, especially when we can use the Shuffling Vaccine.} %
In \cref{tab:ablation}, we compare the model performance by the ablation of Shuffling Vaccine and student Self-Retrospection.
We show that shuffling may fail on specific random states and SR can successfully salvage the student from the failure.
Therefore, we conclude that the Shuffling Vaccine may fail on some triggers and require careful selection using synthetic data.
When the vaccine fails, the student SR can salvage the case by exploring the strong signal of triggers.
Notice that the student SR is less effective if the trigger itself is not strong enough.
For example, BadNet has a relatively low ASR without any defense, while SR can only reduce the ASR to 76.2\%.
This is because the trigger is not the strongest noise-biasing model prediction, and therefore it has a lower chance to be synthesized by the student SR.

\begin{table}[!h]
    \centering
    \small
    \begin{tabular}{cc|cc}
    \toprule
     SV & SR  &  BadNets (grid)  & Trojan WM \\
     \midrule
                  &              &   70.7/87.8                & 82.7/93.9 \\
     $\checkmark$ &              & 67.2/\textbf{0.3}          & 79.0/57.0 \\
                  & $\checkmark$ & 68.3/76.2         & 79.7/44.1 \\
     $\checkmark$ & $\checkmark$ & 68.3/\textbf{0.7} & 78.2/\textbf{22.5} \\  %
     \midrule
     \multicolumn{2}{c}{Clean KD} & 74.6/4.3 & 77.5/11.1  \\
     \bottomrule
    \end{tabular}
    \caption{Ablation study of components on CIFAR-10 with two triggers. Report results as Acc/ASR. To show the failure of shuffling, we disable the selection of shuffling here.}
    \label{tab:ablation}
\end{table}

\begin{figure}[!h]
    \centering
    \includegraphics[width=0.21\textwidth]{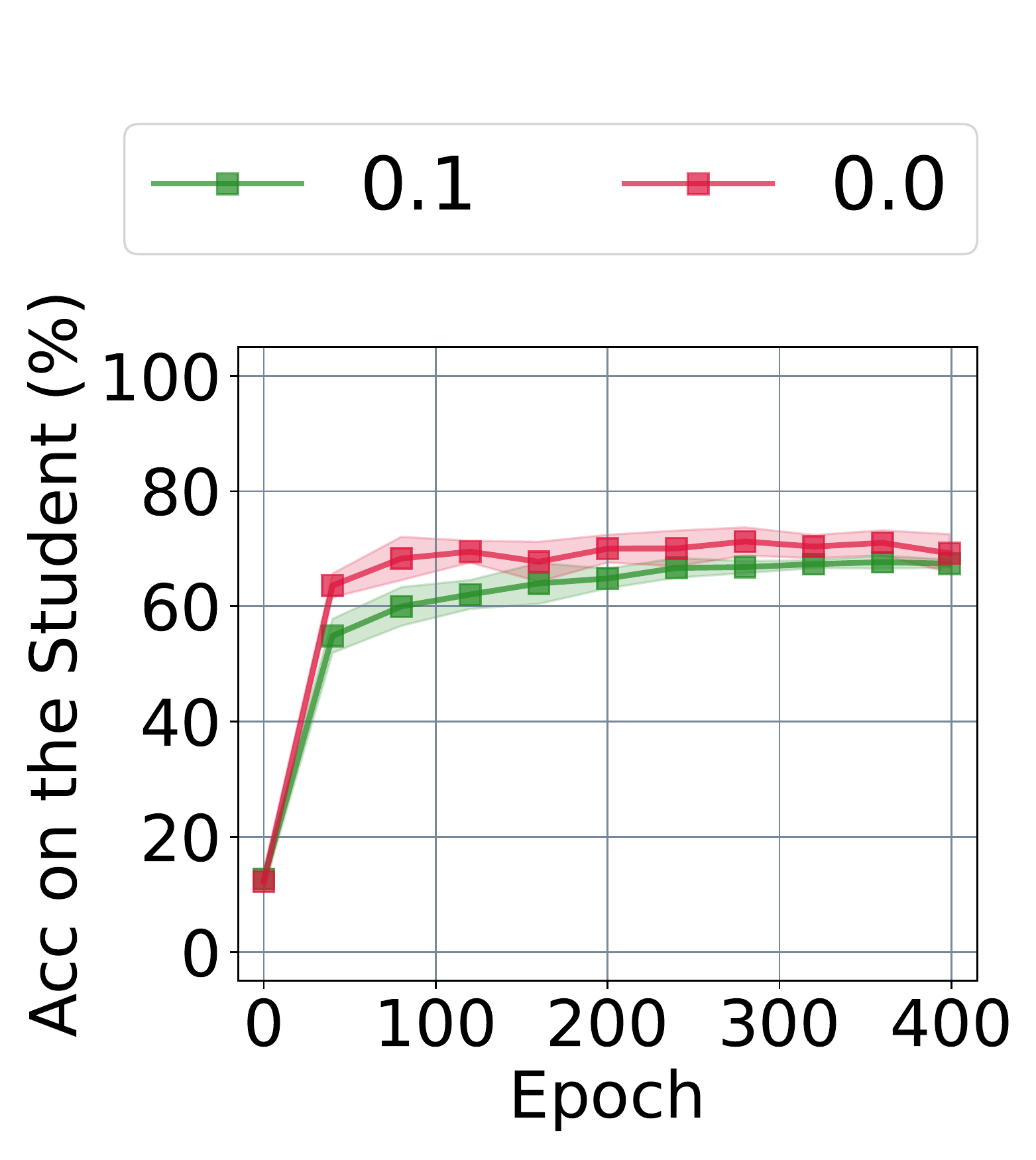}
    \includegraphics[width=0.21\textwidth]{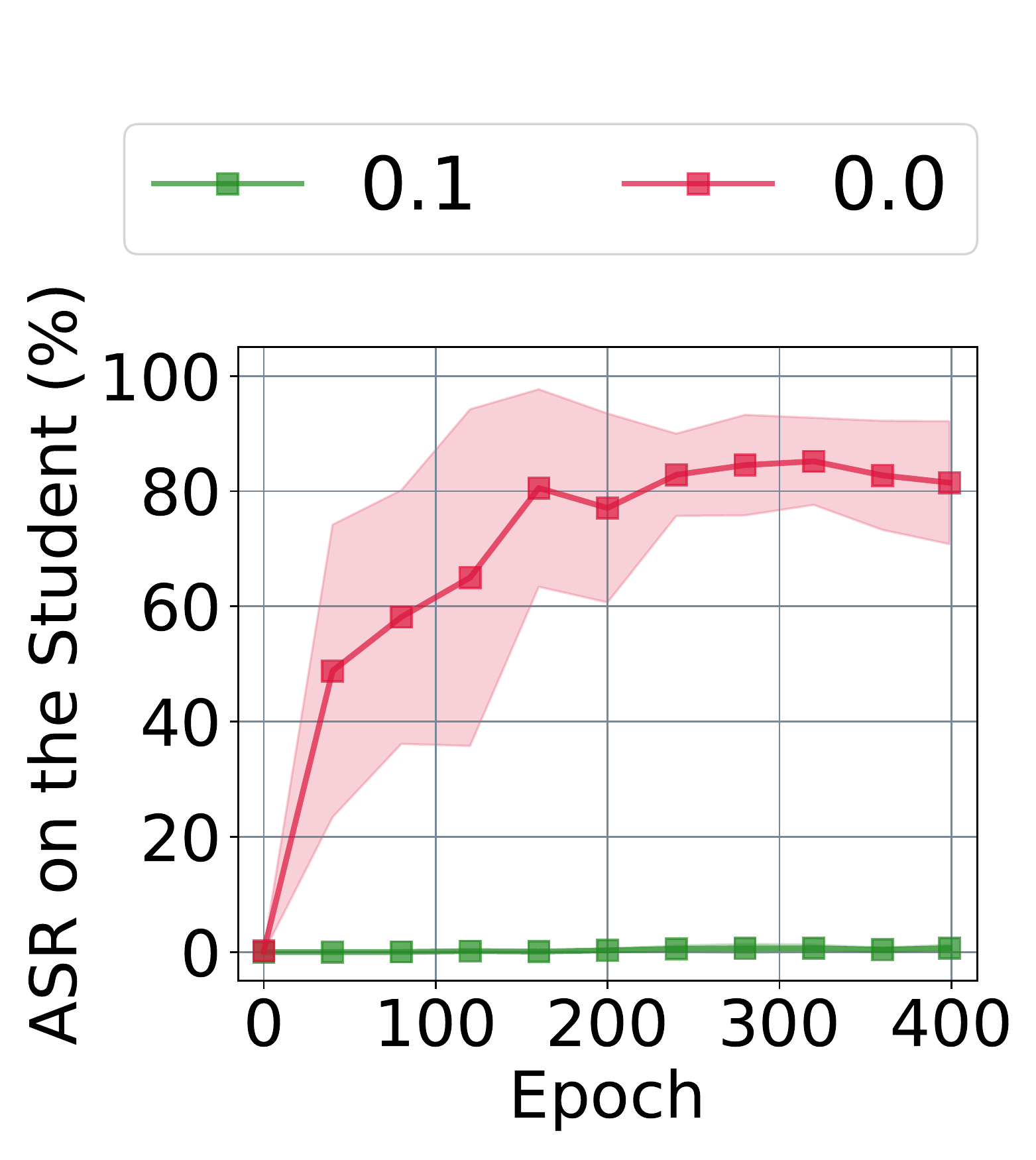}
    \includegraphics[width=0.21\textwidth]{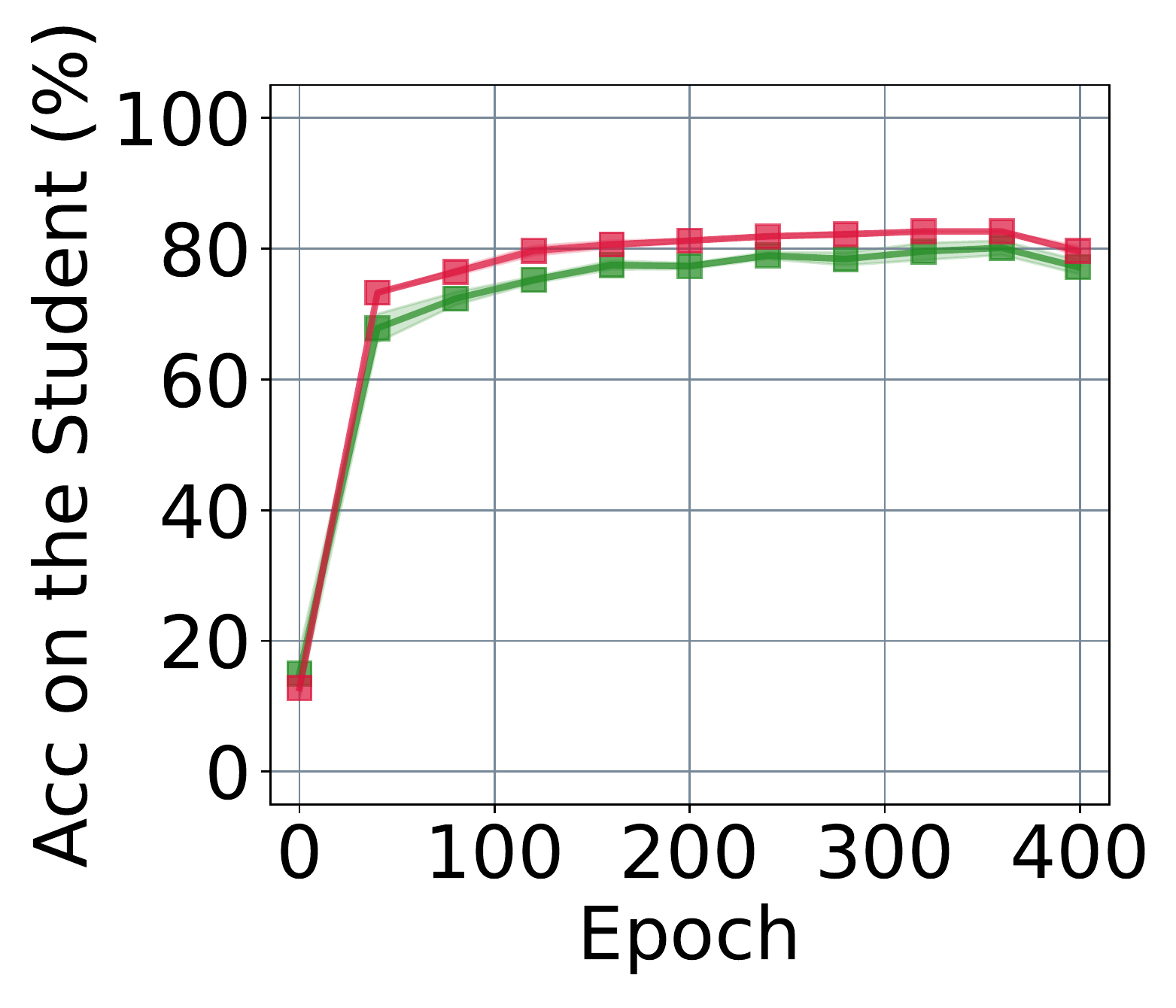}
    \includegraphics[width=0.21\textwidth]{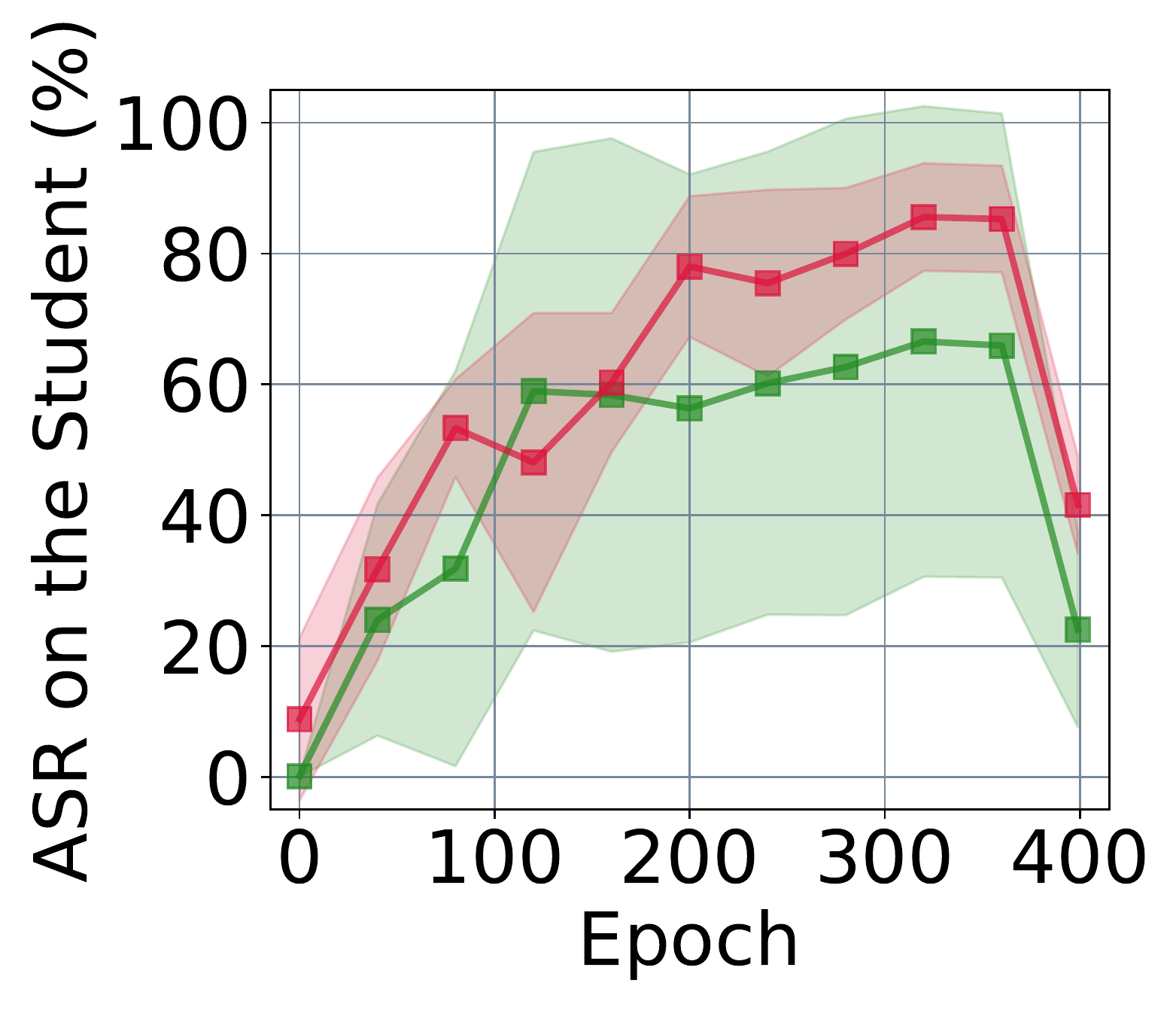}
    \caption{Ablation of the Shuffling Vaccine. The upper two figures are BadNets (grid) and the below two are Trojan WM.}
    \label{fig:defend_acc_asr}
    \vspace{-1.3em}
\end{figure}

\textbf{Shuffling Vaccine in learning.}
In \cref{fig:defend_acc_asr}, we show how the vaccine suppresses the poisons during distillation and how it fails.
To show the failure case of Shuffling Vaccines, we do not make selections here.
When defending against BadNets (grid), the Shuffling Vaccine is able to compromise the ASR to almost 0 at the very beginning, which process almost does not affect the convergence on the clean accuracy.
On the failure of Shuffling Vaccine, the student Self-Retrospection is applied at the last 5 epochs, which effectively suppresses the Trojan WM backdoors.

\section{Conclusion and Discussions}
In this work, we make the first effort to reveal the security risk of data-free KD w.r.t. the untrusted pre-trained models. To mitigate the chance of potential backdoors in the synthetic or OOD data being transferred, we propose ABD, the first plug-in defensive method for data-free KD methods. We empirically demonstrate that ABD can diminish transferred backdoor knowledge while maintaining comparable downstream performances as the vanilla KD.

\textbf{Limitation and broader impact.} 
We believe our findings could be inspiring for the research in making data-free KD secure and the downstream student trustworthy.
As a pilot work, we expect more interesting studies on understanding the poisoning mechanism behind the data-free distillation and defending multiple triggers~\cite{cai2022randomized} and an ensemble of teacher models for federated learning~\cite{zhang2022dense}.
When pre-training data are unavailable, a data-free detection strategy could be desired to determine whether the model is ever poisoned.
Except for distillation, it is also intriguing to study if pre-trained models can be poisonous under robust training~\cite{hong2021federated}.
Beyond the scope of this paper, the \emph{dark side} of our backdoor mitigation could be the risk on backdoor-based Intelligence-Property (IP) protection \cite{jia2020entangled,li2019prove}.
In IP protection, a backdoor is injected for later verifying the ownership of an unauthorized model distillation.
The effectiveness of our mitigation indicates that removing the IP backdoor is possible even without training data.

\section*{Acknowledgments}

This work is supported partially by Sony AI, NSF IIS-2212174 (JZ), IIS-1749940 (JZ), NIH 1RF1AG072449 (JZ), ONR N00014-20-1-2382 (JZ), a gift and a fellowship from the Amazon-VT Initiative.
We also thank anonymous reviewers for providing constructive comments.
In addition, we want to thank Haotao Wang from UT Austin for his valuable discussion when developing the work.

\bibliography{ref}
\bibliographystyle{icml2023}

\clearpage

\appendix

\section{Experimental Details}
\label{sec:app:exp}

In this section, we provide details of hyper-parameters.
To verify shuffling models, we cache 50 batches for ZSKT and 100 batches on OOD as $D_s$.
Shuffling Vaccine is done by randomly changing the order of channels in the last 5 convolutional layers of WideResNet (corresponding to the last stage) and an ensemble of three shuffled models is used.
If SV significantly degrades the clean accuracy, we will restart the distillation without SV.
The Self-Retrospection treatment is done at the last 3 epochs of CMI/OOD, 800 batches of ZSKT.

For ZSKT on GTSRB, we tune the KL temperature until maximizing the student's clean accuracy.
The preferred temperature will be 0.5, and we remove the feature alignment, which yields better accuracy.

For OOD, we use the pre-sliced $10,000$ patches provided by the authors and augment the patches by random CutMix with 100\% probability and $\beta=0.25$ for the Beta-sampling.

For CMI, we directly use the hyper-parameter set provided by the authors for distillation from WRN16-2 to WRN16-1 on CIFAR10.
As commented by the authors, these parameters are very sensitive, and therefore, we do not change them.

All the defense experiments are repeated three times using the seed set $\{0,1,2\}$.
For pre-training backdoored teachers, we use the published codes \citep{wang2022trap}\footnote{\url{https://github.com/VITA-Group/Trap-and-Replace-Backdoor-Defense}}.
Note that the codes do not normalize the input, but we follow the common practice to normalize the CIFAR10 and GTSRB inputs, following~\cite{zeng2021adversarial}.

\section{The Prevalence of the Risk}
\label{sec:app:risk}
\begin{figure}[t]
  \centering
  \includegraphics[width=0.49\linewidth]{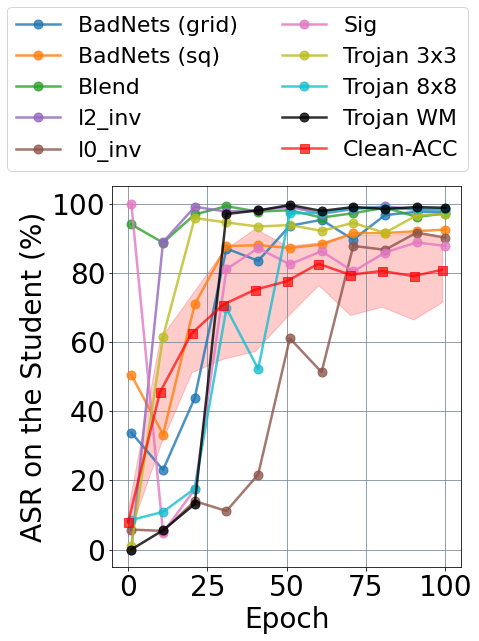}
  \caption{The risk of backdoor infiltrating the student model on the GTSRB-dataset settings (ZSKT as the data-free KD method).
  }
  \label{fig:app:gtsrb}
\end{figure}

We further demonstrate the risk of backdoor infiltrating from the teacher model to the student via data-free KD under a different dataset setting, i.e., using the GTSRB dataset. As shown in Figure \ref{fig:app:gtsrb}, the observation under the GTSRB-dataset setting is consistent with Figure \ref{fig:example_attacks}. Beyond the prevalence of the risk, it seems the dataset with lower sample diversity (compared to CIFAR-10, GTSRB data points are traffic signs taken from almost the same angle, which of lower sample diversity within each class) suffers with a higher risk (all the evaluated attacks are transferred to the student under the GTSRB setting) of backdoor being transferred via data-free KD.
In addition, we examine the transfer from ResNet34 to ResNet18 on PubFig dataset~\cite{kumar2009attribute}.
The ASR can approach $88.6\%$ at the end using trojan\_wm trigger when benign accuracy reaches 87.1\% after 10,000 epochs.

\section{Plausible Understanding of the Security Risk w.r.t. Data-free Settings}

\begin{figure}[!h]
    \centering
    \includegraphics[width=\columnwidth]{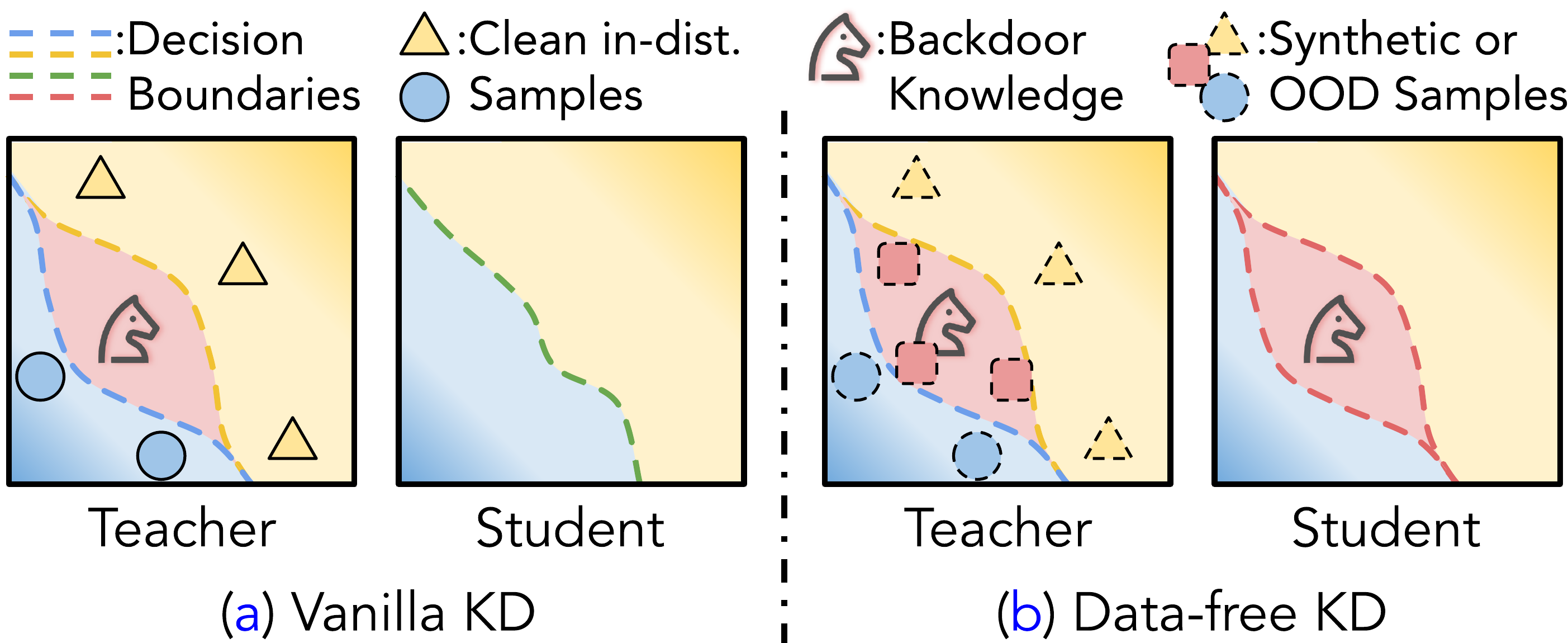}
    \caption{
    The difference between vanilla KD and data-free KD. The use of synthetic and OOD samples may be of low confidence w.r.t. all vicinity classes (\scalebox{0.9}{\colorbox[HTML]{c8ddf1}{\textbf{blue}}} and \scalebox{0.9}{\colorbox[HTML]{ffefbf}{\textbf{yellow}}} region) thus activating the backdoor knowledge from the poison teacher (\scalebox{0.9}{\colorbox[HTML]{F3CECA}{\textbf{red}}} region, which previously cannot be activated via clean in-distribution samples) and leading to the transfer of backdoor knowledge.
    }
    \label{fig:data-free}
\end{figure}

\noindent
\textbf{Why does data-free not lead to poison-free?}
This question is non-trivial since the distillation samples are either generated via an additional generative network or sampled from OOD; see examples from \cref{fig:examples}, which visually do not contain the initial triggers.
As most existing backdoor attacks require poisoning of the training samples, it is unclear the main cause of the transfer of backdoor knowledge under data-free settings.
As the key difference between vanilla KD and data-free KD methods is the participants of synthetic or OOD samples that of lower confidence w.r.t. the output logits of the teacher, one possible reason is that some of these low-confidence points activate similar neurons of the poisoned teacher as the initial poisoned samples, thus leading to the backdoor knowledge being transferred. An intuition of the presumption is depicted in \cref{fig:data-free}.

\begin{figure}[!t]
    \centering
    \includegraphics[width=0.8\columnwidth]{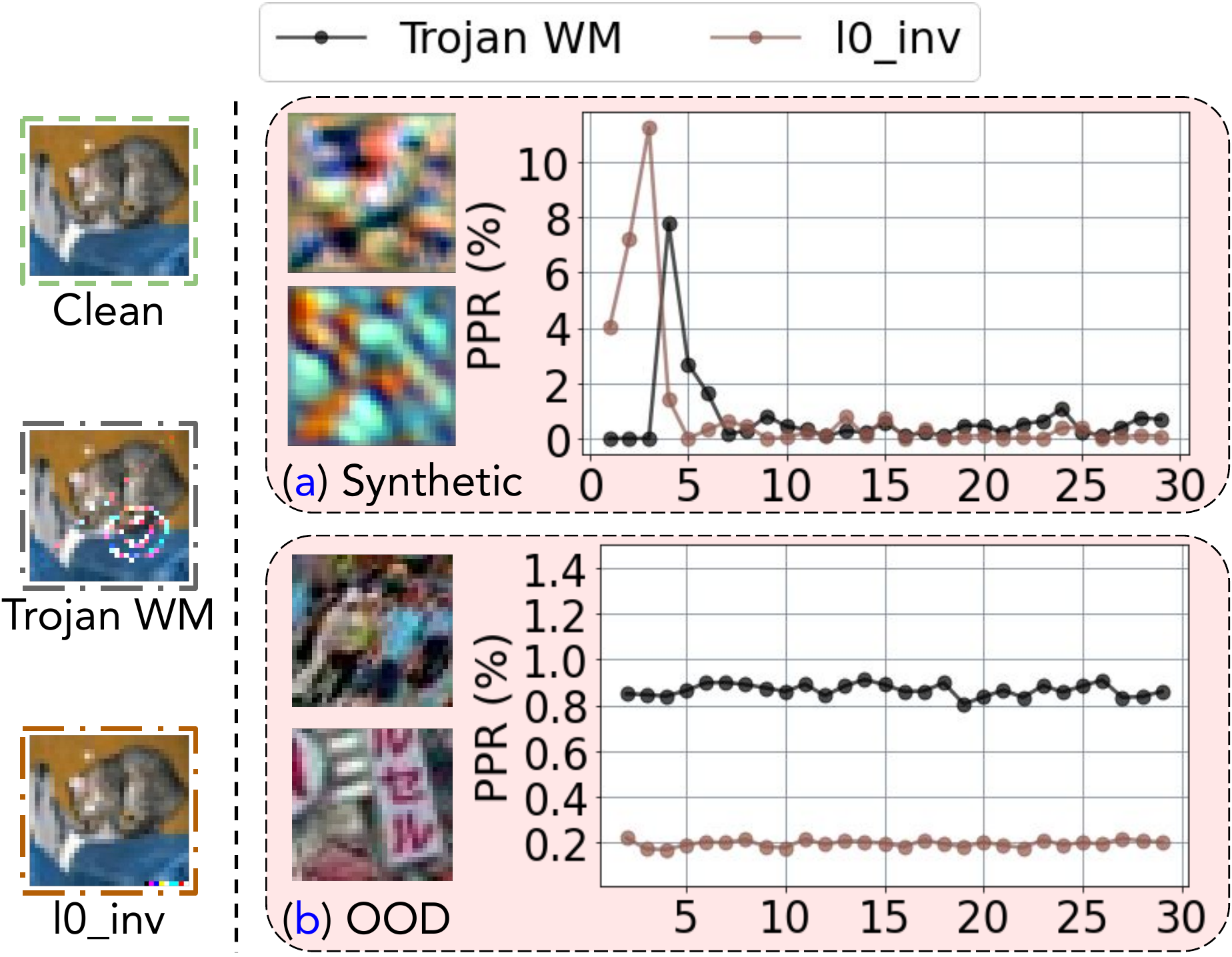}
    \caption{
    An empirical study of whether synthetic and OOD samples directly activate backdoor knowledge. 
The chance of a synthetic data point being classified as a poisoned sample is interpreted as the Plausible Poison Ratio (PPR).
The left-hand side shows the clean and poisoned examples that were used for this experiment. The right-hand side depicts the visual examples and PPR results of KD methods based on (\textcolor{blue}{a}) synthetic or (\textcolor{blue}{b}) OOD samples, respectively. 
    }
    \label{fig:cover-transfer}
\end{figure}

Following this idea, we further explore if there's a direct observation indicating that synthetic samples or OOD can activate backdoor knowledge. We train a logistic regression classifier that takes the teacher model's output logits as the input to see if synthetic and OOD data may activate similar neurons as poisoned samples and how the portion of the sample may affect the transfer of backdoor knowledge. 
To start, we use the test set of CIFAR-10 to obtain 9,000 output logits of poisoned samples (patched with the initial trigger and labeled as 1) and 10,000 output logits of clean samples (labeled as 0).
We then train the logistic regression classifier with the above two categories of logits and apply it to unseen synthetic/OOD samples' output logits from the same epoch and measure the false positive rate of synthetic/OOD data being classified as positioned samples (plausible poison ratio, or PPR).
We depict the analysis of the experiment on the Trojan WM (example of an attack that always infiltrates the student) and l0\_inv (example of an attack that cannot infiltrate the student) in Figure \ref{fig:cover-transfer}. 
The insight on OOD samples is quite clear, where we find OOD samples can activate similar neurons as the initial poisoned samples on success attack case (Trojan WM) with 3 to 4 $\times$ higher PPR than the failed attack (l0\_inv). This, in a way, aligns with our presumption that the backdoor knowledge is activated with input samples being similar to the initial poisoned data. However, the results of Synthetic data is hard to come to the same conclusion. We suspect that the reason why backdoor attacks can infiltrate data-free KD based on synthetic data is more complicated, and we defer it to future work of exploration.

\begin{figure}[ht]
    \centering
    \includegraphics[width=0.4\textwidth]{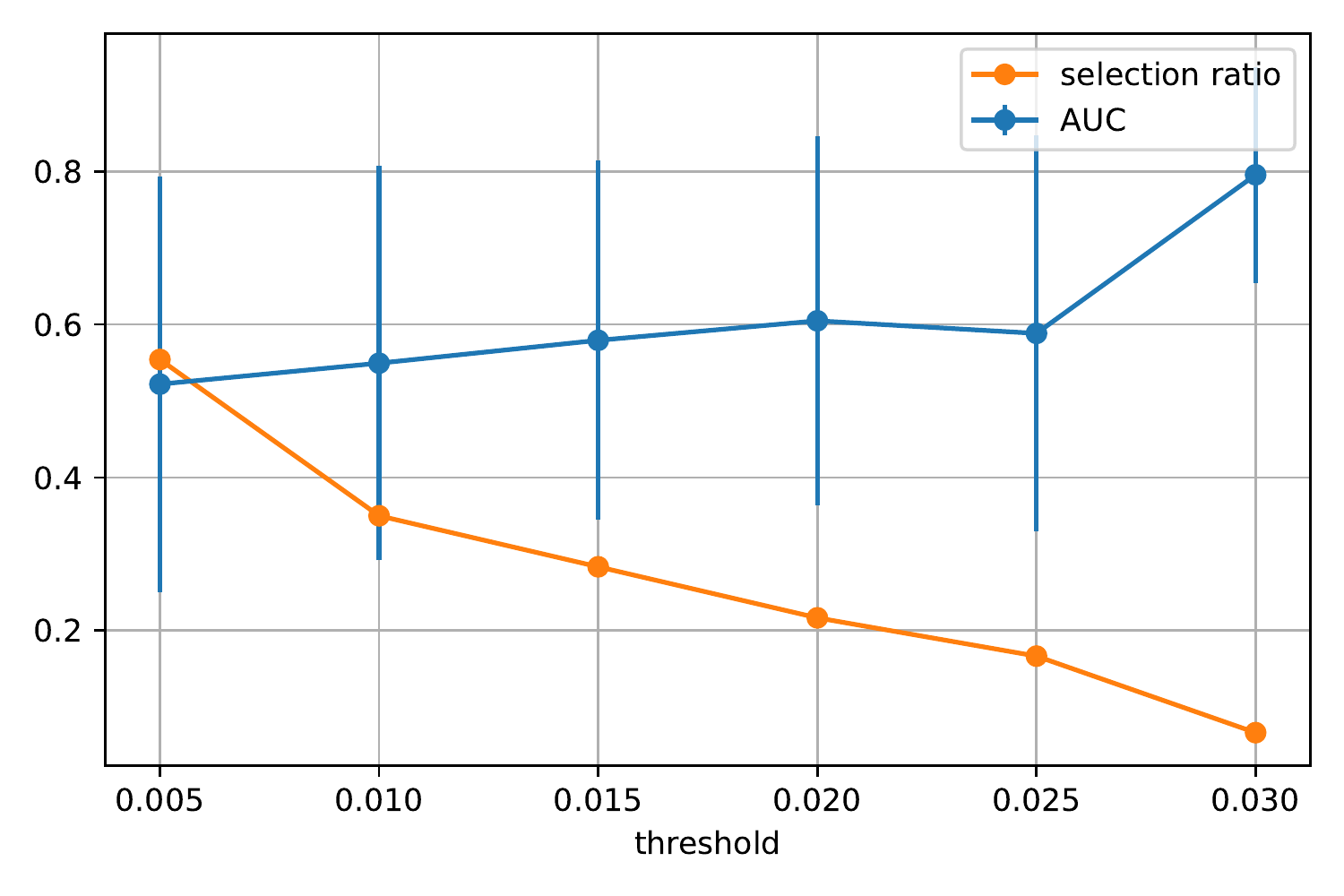}
    \caption{Ablation study on the threshold.}
    \label{fig:ablation_threshold}
\end{figure}

\section{Ablation Study on SV's Threshold}
\label{sec:app:ablation}

In \cref{fig:ablation_threshold}, we conduct an ablation study on the threshold for selecting shuffle models.
We conduct experiments using models with Trojan WM and BadNet (grid). We compute the backdoor detection AUC when varying the threshold. 
A higher AUC is desired for accurately recognizing poison samples.
We also report the ratio of shuffle models selected in the sample set. 
A higher selection ratio means that it is easy to find a desired shuffled model in a limited sample set and therefore is more efficient.
In each case, we sample 20 shuffle models.
As shown in \cref{fig:ablation_threshold}, the threshold 0.02 can strike a balance between high AUC and a reasonable selection ratio. By the selection ratio 0.22, we only need approximatly 5 samples to find an effective shuffle model in expectation, which is smaller than our sample size 8.

\end{document}